\documentclass[a4paper, 10pt, conference]{ieeeconf}

\usepackage{xcolor}
\usepackage{lipsum}
\usepackage{cite}
\usepackage{amsmath,amssymb,amsfonts}
\usepackage{algorithmic}
\usepackage{graphicx}
\usepackage{siunitx}
\usepackage[utf8]{inputenc}
\usepackage{textcomp}
\usepackage{amsmath}
\usepackage{amsfonts}
\usepackage{commath}
\usepackage{float}

\usepackage{epsfig}
\usepackage{epstopdf}
\usepackage{amsmath}
\usepackage[fleqn]{nccmath}
\usepackage[linesnumbered, ruled, vlined]{algorithm2e}

\def\BibTeX{{\rm B\kern-.05em{\sc i\kern-.025em b}\kern-.08em
    T\kern-.1667em\lower.7ex\hbox{E}\kern-.125emX}}

\newcommand*{\rom}[1]{\expandafter\@slowromancap\romannumeral #1@}
\makeatother

\SetCommentSty{mycommfont}
\newcolumntype{P}[1]{>{\centering\arraybackslash}p{#1}}
\newcolumntype{M}[1]{>{\centering\arraybackslash}m{#1}}

\IEEEoverridecommandlockouts

\begin{document}

\raggedbottom
\title{\LARGE \bf Adaptive Force Controller for Contact-Rich Robotic Systems using an Unscented Kalman Filter }


\author{Alexander Schperberg$^{\dagger}$, Yuki Shirai$^{\dagger}$, Xuan Lin$^{\dagger}$, Yusuke Tanaka$^{\dagger}$, and Dennis Hong$^{\dagger}$
\thanks{$^{\dagger}$All authors are with the Robotics and Mechanisms Laboratory, Department of Mechanical and Aerospace Engineering, University of California, Los Angeles, CA, USA 90095 {\tt\small \{aschperberg28, yukishirai4869, maynight, yusuketanaka, dennishong\}@g.ucla.edu}}
}

\maketitle

\global\csname @topnum\endcsname 0
\global\csname @botnum\endcsname 0
\begin{abstract}
In multi-point contact systems, precise force control is crucial for achieving stable and safe interactions between robots and their environment. Thus, we demonstrate an admittance controller with auto-tuning that can be applied for these systems. The controller's objective is to track the target wrench profiles of each contact point while considering the additional torque due to rotational friction. Our admittance controller is adaptive during online operation by using an auto-tuning method that tunes the gains of the controller while following user-specified training objectives. These objectives include facilitating controller stability, such as tracking the wrench profiles as closely as possible, ensuring control outputs are within force limits that minimize slippage, and avoiding configurations that induce kinematic singularity. We demonstrate the robustness of our controller on hardware for both manipulation and locomotion tasks using a multi-limbed climbing robot. 
\end{abstract}


\section{Introduction}
Forces and torques are exchanged in nearly every type of interaction between robots and their environments. From robotic manipulators that perform peg-hole-insertion tasks \cite{peg_hole}, pushing/pulling objects \cite{pushing,push_pull}, to the ground reaction wrenches generated by foot contacts for robotic climbing \cite{risk_aware}. Thus, state-of-the-art algorithms for locomotion and manipulation estimate desired wrench profiles which can achieve stable balancing of an object \cite{stable_grasp} or satisfy kinematic and dynamic constraints \cite{trajOpt}. Additionally, current controllers typically use forces as control variables, optimizing them to either stabilize the Center of Mass trajectory of the robot's base \cite{mpc} or achieve compliance control during manipulation
\cite{compliance_ctrl}. In all the above cases, the success of the robot's motion depends on how well a controller can comply with external perturbation and ideally, track the desired output wrench profiles.

In this paper, our objective is to formulate an auto-calibrating admittance controller for tracking the wrench profile using F/T sensors in a closed-loop fashion. 
The main contribution of this paper is to demonstrate this using an Unscented Kalman Filter (UKF) to track wrench profiles that includes the additional torque due to rotational friction, while considering training objectives that facilitate controller robustness and adaptability during online operation for multi-point contact grippers. 

\begin{figure}[!t]
    \centering
    \includegraphics[width=0.95\columnwidth]{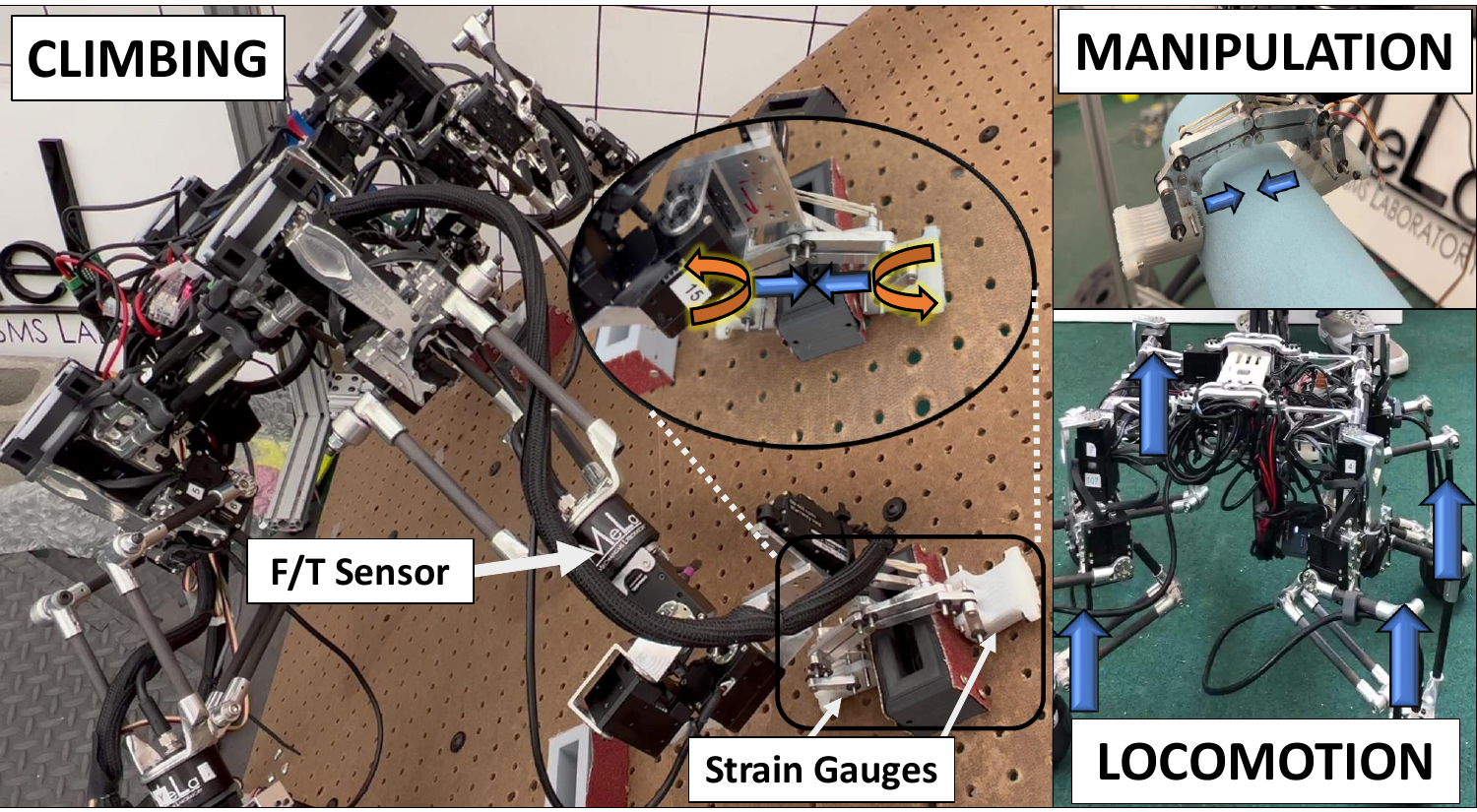}
    \caption{\textbf{Hardware validation tests.} Auto-calibrating admittance control for tracking wrenches for climbing, manipulation, and locomotion tasks. The blue arrows indicate forces, while the orange arrows indicate torques. We also indicate the Force/Torque (F/T) sensor at the wrist, and strain gauges located on the finger, measuring the finger compression force}
    \label{intro_pic}
\end{figure}

Additionally, we validate our controller on hardware (Figure. \ref{intro_pic}) that consists of under-actuated grippers, and demonstrate tracking of reference wrench profiles for various tasks, including wall climbing, grasping of objects, and locomotion. 


\section{Related Works}
Force control strategies are worthwhile pursuits as they have already been shown, compared to ignoring force feedback, to deliver adequate control for diverse tasks, from insertion with tight tolerances to the polishing of non-flat surfaces \cite{polishing}. So far, one area that is relatively less explored is the control of wrenches for dexterous manipulation, which typically involve one or more fingertips that must physically interact with the environment and objects (i.e., multi-finger robots) \cite{compliance_ctrl}. Instead, the research focus in dexterous manipulation has been on object recognition or localization with fully-actuated grippers, where grasping motion is applied through an on/off mechanism, where the gripper is either in a fully open or fully closed configuration \cite{grasp_localization1}. More complex controls for grasping have been considered using tactile feedback and even learning-based approaches through teleoperation \cite{compliance_ctrl2,tactile2}. While these works incorporate wrench information, their goal is to control the relative motion between the gripper and its target object than explicitly track a desired wrench. However, tracking wrenches can be necessary if we have an underactuated gripper or make contact in very compliant environments, as they are difficult to control, and are endowed with the ability to envelop objects of different geometric shapes. Still, control of underactuated grippers has been shown in past work by estimating force but not controlling for them directly in \cite{ballesteros2020proprioceptive} or controlling force directly but without hardware results in \cite{force_sim}. In our work, we will primarily focus on tracking wrench profiles directly using force-feedback with an auto-calibrating admittance controller that can be used not only for single-point contacts, such as legged robots with point feet \cite{stable_grasp}, but also for multi-finger robots, such as manipulators with grippers \cite{ballesteros2020proprioceptive}, on both stiff and compliant environments. Note, that we opt for an admittance force controller because we assume access to F/T sensors, and will use a change in task-space position for control after obtaining these sensor measurements.

The reason we employ a self-calibrating admittance controller instead of using traditional admittance control \cite{impedance} is that the controller can quickly become unstable if no modifications are made. For instance, admittance control requires inverse kinematics, where the position of the end-effector as a control parameter can cause kinematic singularities. Also, task and joint space control require a complete dynamic model of the robot, contributing to the accuracy/robustness dilemma \cite{impedance}. Recently, model error compensation techniques for impedance/admittance control have been shown through adaptive control schemes, although knowledge of the dynamic model is still required. One approach to resolve model-based errors is demonstrated in \cite{yu_simplified_2020}, which modifies the admittance controller by using only the orientation components of the end-effector through relating orientation with joint angles, avoiding the need for inverse kinematics. However, this approach can only be used if the limb's joint space is $\mathbb{R}^{6}$, ours is $\mathbb{R}^{7}$. Additionally, deriving orientation with joint angles may be difficult if the gripper is too complex or underactuated. Other methods that are model-free, such as neural networks, may still result in imprecise control and demands a large number of training sets to achieve robustness under unseen conditions \cite{neural_network_adm}. 

Alternatively, to modify our admittance controller, we use the method of \cite{menner2021automated} (which employs a UKF) over other adaptive or machine-learning methods as it does not require a trial-and-error implementation and can be directly applied to hardware without the need for complex simulation models--which may otherwise be difficult to achieve with a physically complex manipulator/gripper and in uncertain environments. The method also allows the freedom to choose any number of desired training objectives toward stabilizing the controller, such as closely tracking wrench profiles, predicting the spring constant of the environment to update the dynamic model, ensuring control outputs are within limits to minimize slipping, and avoiding kinematic singularity. 

Although other works exist in tracking dynamic forces in uncertain environments, such as \cite{adaptive1,adaptive2}, their experimental setup is relatively simple and they only show results for tracking force and not torque. By using a UKF, we can also more easily consider nonlinearities imposed by the system. However, the main limitation of our approach relative to those works, is that it is more challenging to provide mathematical guarantees on controller stability. For instance, the admittance controller gains must be positive semi-definite to be stable. To circumvent this issue, we add a small modification to \cite{menner2021automated}, by adding a large cost term on the training objective when the sigma points or controller gains sampled by the UKF violate the semi-definite property. We perform a similar operation to avoid kinematic singularities as well. Doing so, the algorithm will not select gains that violate these properties.




\begin{figure*}[!t]
    \centering
    \includegraphics[width=6.5in]{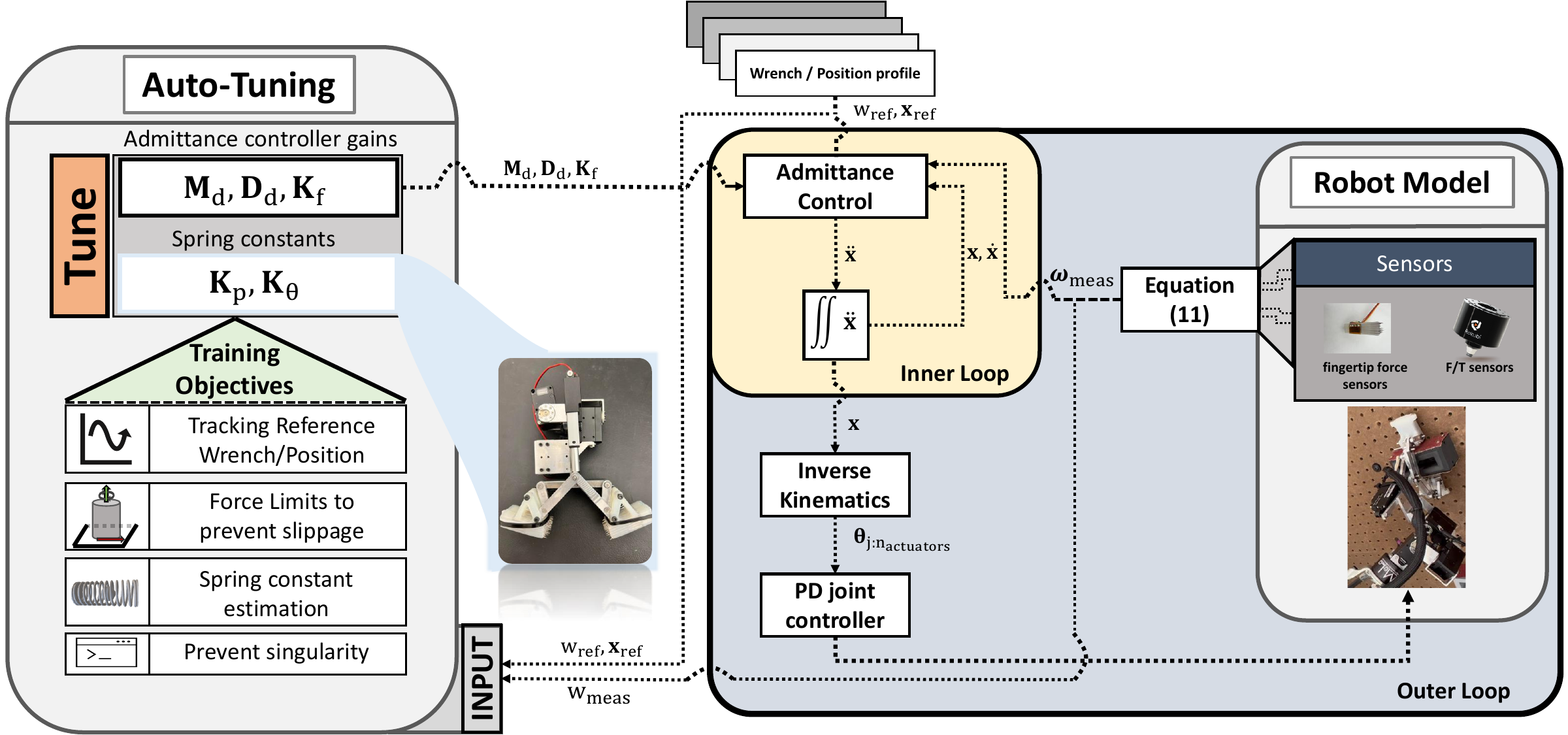}
    \caption{\textbf{Auto-calibrating admittance control framework.} This figure shows the overall control architecture. The admittance controller takes as input the wrench and position profile (represented by $\mathcal{W}_{\rm ref}$ and $\mathbf{x}_{\rm ref}$) or the forces $\mathbf{f}_{\rm ref}$ (if our MPC is used). The current wrenches are received using the relationship provided by equation (11) which requires force/torque sensor information, and also the current control inputs ($\mathbf{x}$ and $\mathbf{\dot{x}}$). The output of the admittance controller is the acceleration ($\mathbf{\ddot{x}}$), where its integration yields the control inputs for the next timestep (considered the inner loop). Since we use position-controlled motors, we use $\mathbf{x}$ as input to our inverse kinematics, which provides the joint motor angles ($\boldsymbol{\theta}_{1:n_{\rm actuators}}$). A PD joint controller is used to track these angles (considered the outer loop). Lastly, the auto-tuning framework takes as input the reference and current wrenches and outputs new gains for the admittance controller ($\mathbf{M}_{d}$, $\mathbf{D}_{d}$, and $\mathbf{K}_{f}$). As the auto-tuning method makes use of the robot model (dependent on spring constants $\mathbf{K}_{p}$, and $\mathbf{K_{\theta}}$), we make use of the current wrenches to continually update these spring constants to a more accurate estimate as part of the auto-tuning process.}  
    \label{fig_methods}
\end{figure*}

\section{Admittance control formulation}
\label{methods:Admittance controller}
The fundamental nature of an admittance controller is that it converts a force input into a motion defined by a change in position. Admittance control may be necessary for robots that have positioned-controlled motors but still need to simulate the properties of compliance when interacting with its environment. For clarity in wording, we will call the end-effector as being the fingertip, and the wrist as being the base of the robot's gripper where the fingertips are attached.
Our admittance control formulation can be described by:
\begin{align}
    \label{eq:admittance_controller_inv}
    \mathbf{\ddot{x}}=\mathbf{M}_{d}^{-1}(-\mathbf{D}_{d}\mathbf{\dot{x}}-\mathbf{K}_{d}(\mathbf{x} - \mathbf{x}_{\rm ref})+ \mathbf{K}_{f}(\mathcal{W}_{\rm meas}-\mathcal{W}_{\rm ref}))
\end{align}
where $\mathcal{W}_{\rm meas}\in\mathbb{R}^{6k}$ is the current wrench and $\mathcal{W}_{\rm ref}\in\mathbb{R}^{6k}$ is the desired reference wrench for fingertip or contact point $k$, 6 represents the $x$, $y$, and $z$ components of force, $\mathbf{f}\in\mathbb{R}^{3k}$, and torque, $\boldsymbol{\tau}\in\mathbb{R}^{3k}$, with $\mathcal{W}=[\mathbf{f}^\top,\boldsymbol{\tau}^\top]^\top$.  $\mathbf{M}_{d}\in\mathbb{R}^{6k \times 6k}$, $\mathbf{D}_{d}\in\mathbb{R}^{6k \times 6k}$, $\mathbf{K}_{d}\in\mathbb{R}^{6k \times 6k}$, are the diagonal gain matrices that can be described as the desired mass, which must be invertible, damping, and spring coefficients respectively. $\mathbf{K}_f\in\mathbb{R}^{6k \times 6k}$ does not have any physical meaning, but may be tuned depending on the desired sensitivity to changes in wrench.  $\mathbf{\dot{x}}$, and $\mathbf{x}$ are the control outputs, where $\mathbf{x}=[\mathbf{p}^\top,\mathbf{\Theta}^\top]^\top$ represents the fingertip position, $\mathbf{p}\in\mathbb{R}^{3k}$, in $x$, $y$ and $z$, and orientation, $\boldsymbol{\Theta}\in\mathbb{R}^{3k}$, in $x$, $y$, and $z$, which are solved by integrating $\mathbf{\ddot{x}}$ using Euler discretization. $\mathbf{x}_{\rm ref}$ is a desired position and orientation from a reference trajectory. Note, that subtracting Euler angles, as required by the $\mathbf{x}-\mathbf{x}_{\rm ref}$ term, we can employ the techniques found in \cite{Bullo_Euler}.

\begin{figure}[!t]
    \centering
    \includegraphics[width=1\columnwidth]{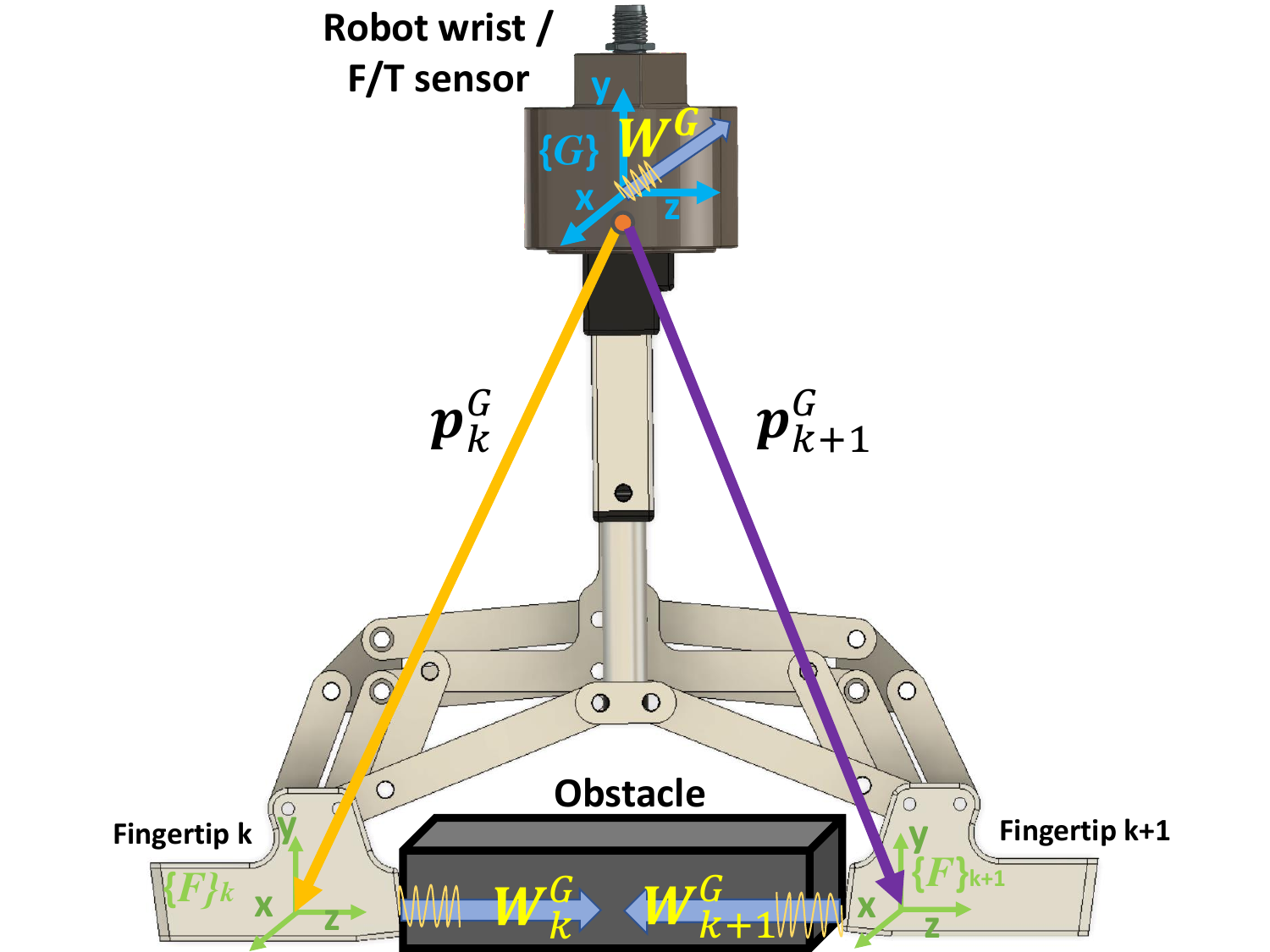}
    \caption{\textbf{Frame definitions.} Here we show the general frame definitions for the variables used in the admittance controller. Our controller will operate in the local gripper frame $\{\mathbf{G}\}$, which is considered the robot's wrist (or the location of the Force/Torque or F/T measurements). The gripper may have $k$ number of end-effectors or fingertips, denoted by frame $\{\mathbf{F}\}$. Fingertip positions, $\mathbf{p}_{k}^{G}$, are all with respect to the gripper frame $\{\mathbf{G}\}$. Similar notations are given to the fingertip wrenches $\mathcal{W}_{k}^{G}$ and wrenches at the wrist $\mathcal{W}^{G}$. Note that to more easily solve for equations, we choose frame $\{\mathbf{F}\}$ to be the same orientation as frame $\{\mathbf{G}\}$.
    }
    \label{frame_def}
\end{figure}

To estimate the current wrench, $\mathcal{W}_{\rm meas}$ (as used in (\ref{eq:admittance_controller_inv})), we must first derive the relationship between wrenches of the robot's wrist and the wrenches of each fingertip of the gripper (frames defined in Figure \ref{frame_def}):

\begin{equation}
\label{torque_wrench}
\mathcal{W}_{k}^{G}=\left(\begin{array}{c}
\mathbf{f}_{k}^{G} \\
\mathbf{p}_{k}^{G} \times \mathbf{f}_{k}^{G}
\end{array}\right)+\left(\begin{array}{c}
\mathbf{0}^{3} \\
\boldsymbol{\tau}_{k}^{G}
\end{array}\right)
\end{equation}
where $\mathcal{W}_{k}^{G}$ is the fingertip wrench of fingertip $k$ relative to the gripper frame $\{\mathbf{G}\}$, $\mathbf{p}_{k}^{G}$ is the position of the fingertip, $\mathbf{f}_{k}^{G}$ is the reaction force, and $\boldsymbol{\tau}_{k}^{G}$ is the additional torque due to patch contact at the fingertip relative to the gripper frame. We assume gripper and fingertip frames are the same, or $\{\mathbf{G}\}$ = $\{\mathbf{F}\}$ in Figure. \ref{frame_def}.

When the fingertips are in contact with the environment, we will assume mathematically that the force/torque at the wrist of the gripper and the sum of forces/torques received from the fingertips are equal. Thus, we have:
\begin{align}
    \label{eq:general_force}
     \mathbf{f}^{G}=\sum_{k=1}^{n_{f}}\mathbf{f}_{k}^{G}
\end{align}
\begin{align}
    \label{eq:general_torque}
    \boldsymbol{\tau}^{G} = \sum_{k=1}^{n_{f}}(\mathbf{p}_{k}^{G} \times \mathbf{f}_{k}^{G}) + \sum_{k=1}^{n_{f}}\boldsymbol{\tau}_{k}^{G} 
\end{align}
where $\mathbf{f}^{G}$ and $\boldsymbol{\tau}^{G}$ are the forces and torques at the wrist, and $n_{f}$ is the number of fingertips. In our case, we have a two-finger gripper ($n_{f}=2$), which would yield a total of 6 equations and potentially 18 variables. 

In this paper, we assume that when the end-effector comes into contact with its environment, some non-negligible amount of deformation occurs between the end-effector material and its environment, resulting in a contact $\lq$patch', which we assume is circular. Thus, the effect of deformation when adapted to a typical rigid body point contact model is an additional torque ($\boldsymbol{\tau}^{n}\in\mathbb{R}^{3}$) that acts about the contact normal \cite{kao_contact_2008}, with Coulomb friction acting on each point of the patch. Using the derivation as done in \cite{kao_contact_2008}, the norm of this torque is bounded by $|\boldsymbol{\tau}^{n}| \leq \boldsymbol{\lambda} \mathbf{f}^{n}$, where $\mathbf{f}^{n}\in\mathbb{R}^{3}$ is the normal force and $\boldsymbol{\lambda} > 0$ is the friction coefficient due to rotation. Throughout this paper, we will assume that the normal force is in the same direction as the $z$ component of fingertip $k$ in the gripper frame $\{\mathbf{G}\}$ (see Figure. \ref{frame_def}) or ${\tau}^{n}$=$\tau_{k,z}^{G}$ and ${f}^{n}=f_{k,z}^{G}$ \cite{kao_contact_2008}.

We also assume access to force/torque sensors at the wrist and 1-axis strain gauges on each fingertip measuring the $z$ component of the force, which we approximate is along the same direction as the surface normal. Thus, we can directly measure $f_{x}^{G},f_{y}^{G},f_{1,z}^{G},f_{2,z}^{G},\tau_{x}^{G},\tau_{y}^{G},$ and $\tau_{z}^{G}$. A limitation of this work is that we don't directly measure all components of the wrench for each fingertip, and thus, need to estimate these wrenches and also impose additional constraints for which components we can independently control. For one, we have an underactuated gripper, thus, in our case, we cannot independently control both fingertip positions in the $x$, and $y$ directions but can for $z$ using the robot's arm and by closing/opening the gripper. Second, we impose constraints in our controller, where we assume control of each rotational axis for torque for both fingertips simultaneously. In other words, taking the above gripper configuration and constraints into account, we have that $f_{1,x}^{G}=f_{2,x}^{G}$, $f_{1,y}^{G}=f_{2,y}^{G}$, $\tau_{1,x}^{G}=\tau_{2,x}^{G}$, $\tau_{1,y}^{G}=\tau_{2,y}^{G}$, and  $\tau_{1,z}^{G}=\tau_{2,z}^{G}$, where only $f_{1,z}^{G}$ and $f_{2,z}^{G}$ can be independently controlled per fingertip--see Sec. \ref{adm_implementation} for details.  
Taking the above simplification and sensor readings into account, we reduce equations \eqref{eq:general_force} and \eqref{eq:general_torque} into the following $\mathbf{A}\mathbf{u}=\mathbf{b}$ form used to solve for $\mathbf{u}$:
\begin{align}
\label{A_matrix}
\begin{split}
& \mathbf{A}=[\mathbf{A}^{1},\mathbf{A}^{2},\mathbf{A}^{3},\mathbf{A}^{4},\mathbf{A}^{5},\mathbf{A}^{6},\mathbf{A}^{7}] \\
& \mathbf{A}^{1} = [2,0,0,0,0,p_{1,z}^{G}+p_{2,z}^{G},-p_{1,y}^{G}-p_{2,y}^{G}]^{\top} \\
& \mathbf{A}^{2} = [0,2,0,0,-p_{1,z}^{G}-p_{2,z}^{G},0,p_{1,x}^{G}+p_{2,x}^{G}]^{\top} \\
& \mathbf{A}^{3} = [0,0,1,0,p_{1,y}^{G},-p_{1,x}^{G},0]^{\top} \\
& \mathbf{A}^{4} = [0,0,0,1,p_{2,y}^{G},-p_{2,x}^{G},0]^{\top} \\
& \mathbf{A}^{5} = [0,0,0,0,2,0,0]^{\top} \\
& \mathbf{A}^{6} = [0,0,0,0,0,2,0]^{\top} \\
& \mathbf{A}^{7} = [0,0,0,0,0,0,2]^{\top} \\
& \mathbf{u} = [f_{1,2,x}^{G},f_{1,2,y}^{G},f_{1,z}^{G},f_{2,z}^{G},\tau_{1,2,x}^{G},\tau_{1,2,y}^{G},\tau_{1,2,z}^{G}]^{\top} \\
& \mathbf{b}=[f_{x}^{G},f_{y}^{G},f_{1,z}^{G},f_{2,z}^{G},\tau_{x}^{G},\tau_{y}^{G},\tau_{z}^{G}]^{\top}
\end{split}
\end{align}
where $f^{G}_{1,2,x}$, $f^{G}_{1,2,y}$, $\tau_{1,2,x}^{G}$, $\tau_{1,2,y}^{G}$, $\tau_{1,2,z}^{G}$ implies that the force and torque of fingertips 1 and 2 are assumed the same during control. Because the determinant$(\mathbf{A})=32$ and the matrix rank is 7, $\mathbf{A}$ can be inverted for all cases. Note that $\mathbf{u}$ is used for $\mathcal{W}_{\rm meas}$ in \eqref{eq:admittance_controller_inv}, and $\mathbf{b}$ consists of the direct measurements from our sensors.  

\subsection{Controlling for grasping force and normal force offsets}
\label{adm_implementation}
Although we can control $f^{G}_{1,2,x}$, $f^{G}_{1,2,y}$, $\tau_{1,2,x}^{G}$, $\tau_{1,2,y}^{G}$, $\tau_{1,2,z}^{G}$ by using (\ref{A_matrix}) directly, where the control output from (\ref{eq:admittance_controller_inv}) assumes force/torques are equal and in the same direction for both fingertips, controlling $f_{1,z}^{G}$ and $f_{2,z}^{G}$ requires extra care as we have the freedom to control these force components independently. To control for these independent normal forces while considering that both fingertips still move either $\lq$inwards' and the gripper is closing, or $\lq$outwards' and the gripper is opening, one solution is to first control for a $\lq$grasping' force, or the amount of normal force exerted on an object, and then separately control for an offset between this grasping force and desired fingertip force. For example, let $f_{1,z,\rm ref}^{G}$, $f_{2,z,\rm ref}^{G}$ be the reference force for fingertips 1 and 2, and $f_{1,z}^{G}$, $f_{2,z}^{G}$ be the measured force for fingertips 1 and 2 from our strain gauges. We then define the reference grasping force as the magnitude $f_{\rm grasp, ref}^{G}=\frac{|f_{1,z, \rm ref}^{G}|+|f_{2,z,\rm ref}^{G}|}{2}$. Note, that the measured grasping force is achieved with this same definition but using $f_{1,z}^{G}$ and $f_{2,z}^{G}$ instead. We then use \eqref{eq:admittance_controller_inv} but with the previous definition of grasping force, where the output $\mathbf{{x}}$ constitutes the fingertip 1 and 2 positions for $z$, or $p_{1,z}^{G}$ and $p_{2,z}^{G}$, where both fingertip positions change with the same magnitude but in opposite directions according to frame $\{\mathbf{G}\}$. However, note that some offset force may exist if $f_{1,z}^{G}$ $\neq$ $f_{2,z}^{G}$ or we desire different reference fingertip forces. To account for this offset force, we let the reference offset force be $f_{z,\rm ref}^{G}=f_{1,z,\rm ref}^{G} - f_{\rm grasp,ref}^{G}$, where the measured force is received from the force/torque sensors at the wrist (or $f_{z}^{G}$). For this case, the control output $\mathbf{x}$ will move both fingertip positions ($p_{1,z}^{G}$ and $p_{2,z}^{G}$) with the same magnitude and same direction. In other words, the robot moves its arm to compensate for offsets between fingertip normal forces as defined by $\{\mathbf{G}\}$. While we can essentially think of having 3 admittance controllers, one for controlling all fingertip components except for $f_{1,z}$, and $f_{2,z}$ one for grasping force, and another for the offset in fingertip normal force, for simplicity, we assume the same set of gains across each. In other words, $\mathbf{K}_f$ for controlling the offset normal force will be the same as $\mathbf{K}_{f}$ for controlling the grasping force.

\section{Auto-tuning formulation}
\label{methods:adaptive}
In this paper, we use a UKF \cite{menner2021automated} to tune the control parameters of an admittance controller to track reference fingertip wrenches (Sec. \ref{reference_trajectory}) while following desired properties such as updating the spring constant of fingertip force and orientation to increase model accuracy (Sec. \ref{spring_eval}), ensuring output wrenches that do not cause slipping motion (Sec. \ref{force_eval}), and are within kinematic boundaries to prevent singularity configurations (Sec. \ref{kinem_eval}). While the method was used to calibrate controller gains for moving a vehicle, we apply it for the first time here for dexterous manipulation tasks, with modification on the training objective through use of large costs to avoid instability.

The goal of the auto-tuning method is to calibrate control parameters of a generic controller $\mathbf{x}_{k,t}^{G}= \boldsymbol{\kappa}_{\theta}(\mathcal{W}_{k,t}^{G})$, based on the state, $\mathcal{W}_{k,t}^{G}$ or wrench of fingertip $k$ at timestep $t$, the control input, $\mathbf{x}_{k,t}^{G}$, where $\mathbf{x}_{k,t}^{G}=[\mathbf{p}_{k,t}^{G\top},\mathbf{\Theta}_{k,t}^{G\top}]^\top$, which is the position and orientation of fingertip $k$ at timestep $t$, sensor measurements, $\mathcal{W}_{k,t,\rm meas}^{G}$ or current actual wrench of fingertip $k$ at timestep $t$ using \eqref{A_matrix}, and training objectives. The control parameters are represented by $\boldsymbol{\theta}_{k,t}$, which consists of the diagonal gains of the admittance controller, $\mathbf{M}_{d}$, $\mathbf{D}_{d}$, and ${\mathbf{K}_{f}}$ and spring constants, $\mathbf{K}_{p}$ and $\mathbf{K}_{\theta}$ from \eqref{eq:force_model} and \eqref{eq:torque_model}.
The control parameters are tuned online during operation from timestep $t-1$ to $t$, although we can also perform this method episodically from timestep $t-N$ to $t$, where $N$ is number of past timesteps with the control law:
\begin{align}
\label{eq:learning_law}
    \boldsymbol{\theta}_{k,t} &=\boldsymbol{\theta}_{k,t-1} + \mathbf{K}_{k,t}\left(\mathbf{y}_{k,t}^{\rm des} - \mathbf{h}(\boldsymbol{\theta}_{k,t})\right)
\end{align}
where $\mathbf{K}_{k,t}$ is the Kalman gain (described in \eqref{eq:UKF}), $\mathbf{y}_{k,t}^{\rm des}$ signifies the desired or nominal values, and $\mathbf{h}(\boldsymbol{\theta}_{k,t})$ is the evaluation function, $\mathbf{r}$, which may be composed of various user-defined training objectives--see Sec. \ref{training_objective}:
\begin{align*}
    \mathbf{h}(\boldsymbol{\theta}_{k,t}) &= \mathbf{r}(\mathcal{W}_{k,t-1}^{G},\boldsymbol{\kappa}_{\theta}(\mathcal{W}_{k,t-1}^{G}))
\end{align*}
Thus, the control objective is specified by minimizing the difference between $\mathbf{y}_{k,t}^{\rm des}$ and $\mathbf{h}(\boldsymbol{\theta}_{k,t})$, where the dynamical system satisfies all specifications when $\mathbf{y}_{k,t}^{\rm des}=\mathbf{h}(\boldsymbol{\theta}_{k,t})$. Note, that the proposed method is model-based, thus:
\begin{align}
    \label{eq:sys_dyn}
    \mathcal{W}_{k,t}^{G}=\mathbf{Dyn}(\mathcal{W}_{k,t-1}^{G},\mathbf{x}_{k,t-1}^{G},\mathcal{W}_{k,t-1,\rm {meas}}^{G})-\mathcal{\hat{W}}_{k,t}^{G}
\end{align}
where $\mathbf{Dyn}(\mathcal{W}_{k,t-1}^{G},\mathbf{x}_{k,t-1}^{G},\mathcal{W}_{k,t-1,\rm meas}^{G})$ is the dynamic model with its input state being the estimated wrench, $\mathcal{W}_{k,t-1}^{G}$, of fingertip $k$ at timestep $t-1$, the control input as the fingertip position/orientation, $\mathbf{x}_{k,t-1}^{G}$, of fingertip $k$ at timestep $t-1$, and actual wrench, $\mathcal{W}_{k,t-1,\rm {meas}}^{G}$, of fingertip $k$ at timestep $t-1$. Because the auto-tuning method is performed recursively online after measurements are received, we can calculate the process noise to be: 
\begin{align}
    \label{eq:process_noise}
    \mathcal{\hat{W}}_{k,t}^{G}=\mathbf{Dyn}(\mathcal{W}_{k,t-1}^{G},\mathbf{x}_{k,t-1}^{G},\mathcal{W}_{k,t-1,\rm {meas}}^{G})-\mathcal{W}_{k,t,\rm meas}^{G}
\end{align}
However, to calculate $\mathcal{\hat{W}}_{k,t}^{G}$ we still need to formulate a model of our system, i.e., estimate or predict the value of $\mathbf{Dyn}(\mathcal{W}_{k,t-1}^{G},\mathbf{x}_{k,t-1}^{G},\mathcal{W}_{k,t-1,\rm {meas}}^{G})$ using measurements at timestep $t-1$. To do so, we first use the control output $\ddot{\mathbf{x}}$ from $\eqref{eq:admittance_controller_inv}$ at timestep $t-1$ and integrate to get fingertip position/orientation at $t$ or $\mathbf{x}_{k,t}^{G}$. We then model the fingertip forces at timestep $t$ as a virtual spring/mass system: 
\begin{align}
    \label{eq:force_model}
    \mathbf{f}_{k,t}^{G}=\mathbf{K}_{p}(\Delta \mathbf{p}_{k,t}^{G})
\end{align}
where $\mathbf{K}_{p} \in\mathbb{R}^{3}$ is a spring constant for each of the $x$, $y$, and $z$ displacements of fingertip positions. Note, that $\Delta \mathbf{p}_{k,t}^{G}=\mathbf{p}_{k,t}^{G}-\mathbf{p}_{k,t-1}^{G}$, where $\mathbf{p}_{k,t}^{G}$ is the control output at timestep $t$ and $\mathbf{p}_{k,t-1}^{G}$ is the control output at timestep $t-1$ for fingertip position. Applying the same principle for modeling fingertip forces but now for torques (with a spring constant $\mathbf{K}_{\Theta} \in\mathbb{R}^{3}$):
\begin{align}
\label{eq:torque_model}
\boldsymbol{\tau}^{G}_{k,t}=\mathbf{K}_{\Theta}(\Delta \boldsymbol{\Theta}_{k,t}^{G})
\end{align}
We can use \eqref{torque_wrench}, \eqref{eq:force_model}, and \eqref{eq:torque_model} to solve for the dynamic model through the following relationship:
\begin{align*}
\label{torque_wrench_propagate}
\mathbf{Dyn}(\mathcal{W}_{k,t-1}^{G},\mathbf{x}_{k,t-1}^{G},\mathcal{W}_{k,t-1,meas}^{G}) = \\ \left(\begin{array}{c}
\mathbf{K}_{p}(\Delta \mathbf{p}_{k,t}^{G}) \\
\mathbf{p}_{k,t}^{G} \times \mathbf{K}_{p}(\Delta \mathbf{p}_{k,t}^{G})
\end{array}\right)+\left(\begin{array}{c}
\mathbf{0}^{3 \times 1} \\
\mathbf{K}_{\Theta}(\Delta \boldsymbol{\Theta}_{k,t}^{G})
\end{array}\right)
\end{align*}
Essentially, the system evolution is simulated from timestep $t-1$ to $t$ using different sets of control parameters $\boldsymbol{\theta}_{k,t}$ or sigma points (see below). 
\allowdisplaybreaks
The Kalman gain is found through a UKF formulation, where $\mathbf{y}_{k,t}^{\rm des}$ is an array of desired values rather than sensor measurements. For further technical details, the reader is directed to \cite{menner2021automated}, here, we will simply state the tuning law for the parameters in~\eqref{eq:learning_law}, which uses the Kalman gain $\mathbf{K}_{k,t}$ found through the following equations: 
\begin{subequations}
\label{eq:UKF}
\begin{align}
    \mathbf{K}_{k,t}
    &=
    \mathbf{C}^{sz}_{k,t}\mathbf{S}_{k,t}^{-1}
    \\
    \mathbf{S}_{k,t} 
    &=  
    \textstyle
    {\rm \mathbf{C}}_v + \sum_{i=0}^{2L}\mathbf{w}^{c,i}(\mathbf{y}_{k,t}^i-\boldsymbol{\hat y}_{k,t})(\mathbf{y}_{k,t}^i-\boldsymbol{\hat y}_{k,t})^{\top}
    \\
    \mathbf{C}_{k,t}^{sz}
    &=
    \textstyle
    \sum_{i=0}^{2L}\mathbf{w}^{c,i}(\boldsymbol{\theta}_{k,t}^i-\boldsymbol{\hat \theta}_{k,t})(\mathbf{y}_{k,t}^i-\boldsymbol{\hat y}_{k,t})^{\top}
    \\
    \boldsymbol{\hat y}_{k,t}
    & = 
    \textstyle
    \sum_{i=0}^{2L}\mathbf{w}^{a,i} \mathbf{y}_{k,t}^i
    \\
    \label{eq:sigmapoints_h}
    \mathbf{y}_{k,t}^i
    & =
    \mathbf{h}(\boldsymbol{\theta}_{k,t}^i)
    \\
    \mathbf{P}_{k,t|t-1} 
    &=
    \textstyle
    {\rm \mathbf{C}}_\theta + \sum_{i=0}^{2L}\mathbf{w}^{c,i}(\boldsymbol{\theta}_{k,t}^i-\boldsymbol{\hat \theta}_{k,t})(\boldsymbol{\theta}_{k,t}^i-\boldsymbol{\hat \theta}_{k,t})^{\top}
    \\
    \boldsymbol{\hat \theta}_{k,t}
    & = 
    \textstyle
    \sum_{i=0}^{2L}\mathbf{w}^{a,i} \boldsymbol{\theta}_{k,t}^i
    \\
    \mathbf{P}_{k,t|t} 
    &=
    \mathbf{P}_{k,t|t-1} 
    - \mathbf{K}_{k,t} \mathbf{S}_{k,t} \mathbf{K}_{k,t}^{\top}
\end{align}
\end{subequations}
where $\boldsymbol{\theta}_{k,t}^i$ with $i\!=\!0,...,2L$ are the sigma points, $\mathbf{w}^{c,i}$ and $\mathbf{w}^{a,i}$ are the weights of the sigma points, $\mathbf{C}_{k,t}^{sz}$ is the cross-covariance matrix, $\mathbf{S}_{k,t}$ is the innovation covariance, and $\mathbf{P}_{k,t|t}$ is the estimate covariance. The weights $\mathbf{w}^{a,i}$ are user-defined and in this paper assume identical weight definition as described in \textit{remark 7} of \cite{menner2021automated}. Lastly, note that the covariance matrix $\mathbf{C}_{\theta}$ which is initialized by the user, defines the aggressiveness of the auto-tuning, while $\mathbf{C}_{v}$ defines the $\lq$weight' given to the components of $\mathbf{y}_{k,t}^{\rm des}$ \cite{menner2021automated}. 
\begin{figure*}[!t]
    \centering
    \includegraphics[width=6.5in]{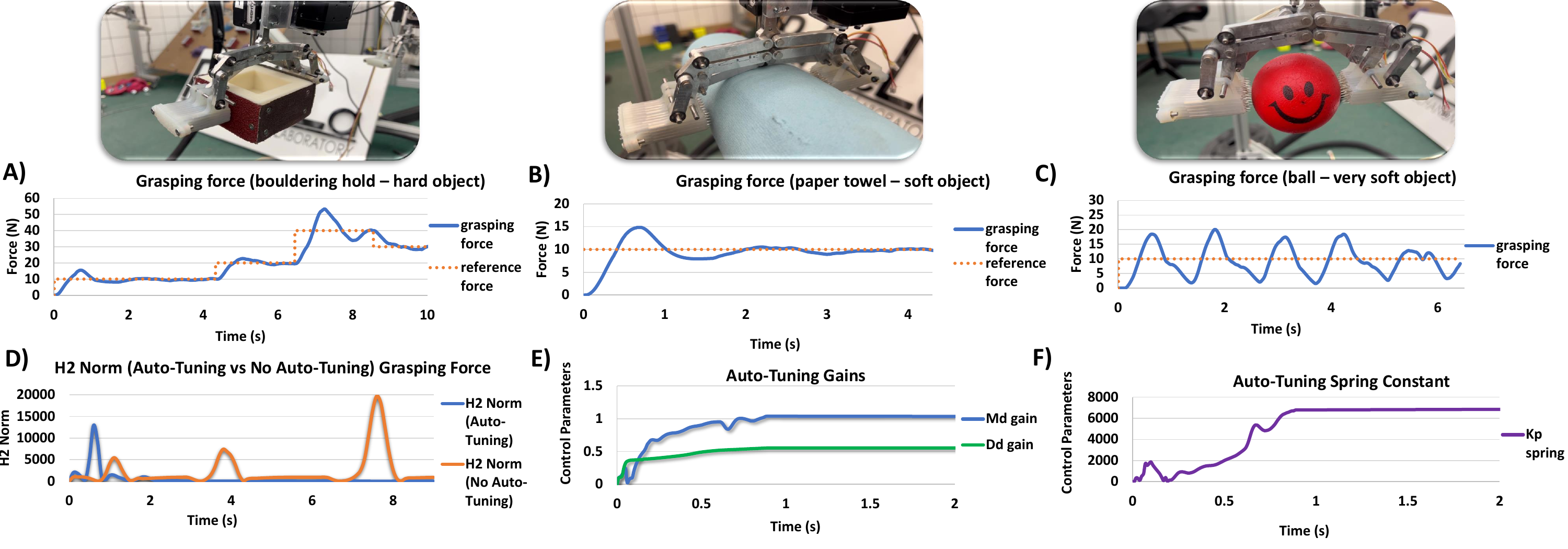}
    \caption{\textbf{Results: Tracking Grasping Force for Manipulation.} ($\mathbf{A}$)-($\mathbf{C}$) shows the results for tracking the grasping force. Note, we do not show the results of auto-tuning versus no auto-tuning for ($\mathbf{A}$)-($\mathbf{C}$) because without auto-tuning, the values can quickly diverge, see ($\mathbf{D}$). ($\mathbf{D}$) is the H2 Norm between auto-tuning (blue) and not auto-tuning (orange) the gains during operation (using the object shown in ($\mathbf{A}$). ($\mathbf{E}$)-($\mathbf{F}$) exemplify the tuning of the $M_{d}$, $K_{d}$ gains of the admittance controller and $K_{p}$ gain of the spring term of the model described in equation \eqref{eq:force_model} (note for this test we only tune the z-component of the fingertip force).}
    \label{results_1}
\end{figure*}
\begin{figure*}[!t]
    \centering
    \includegraphics[width=6.5in]{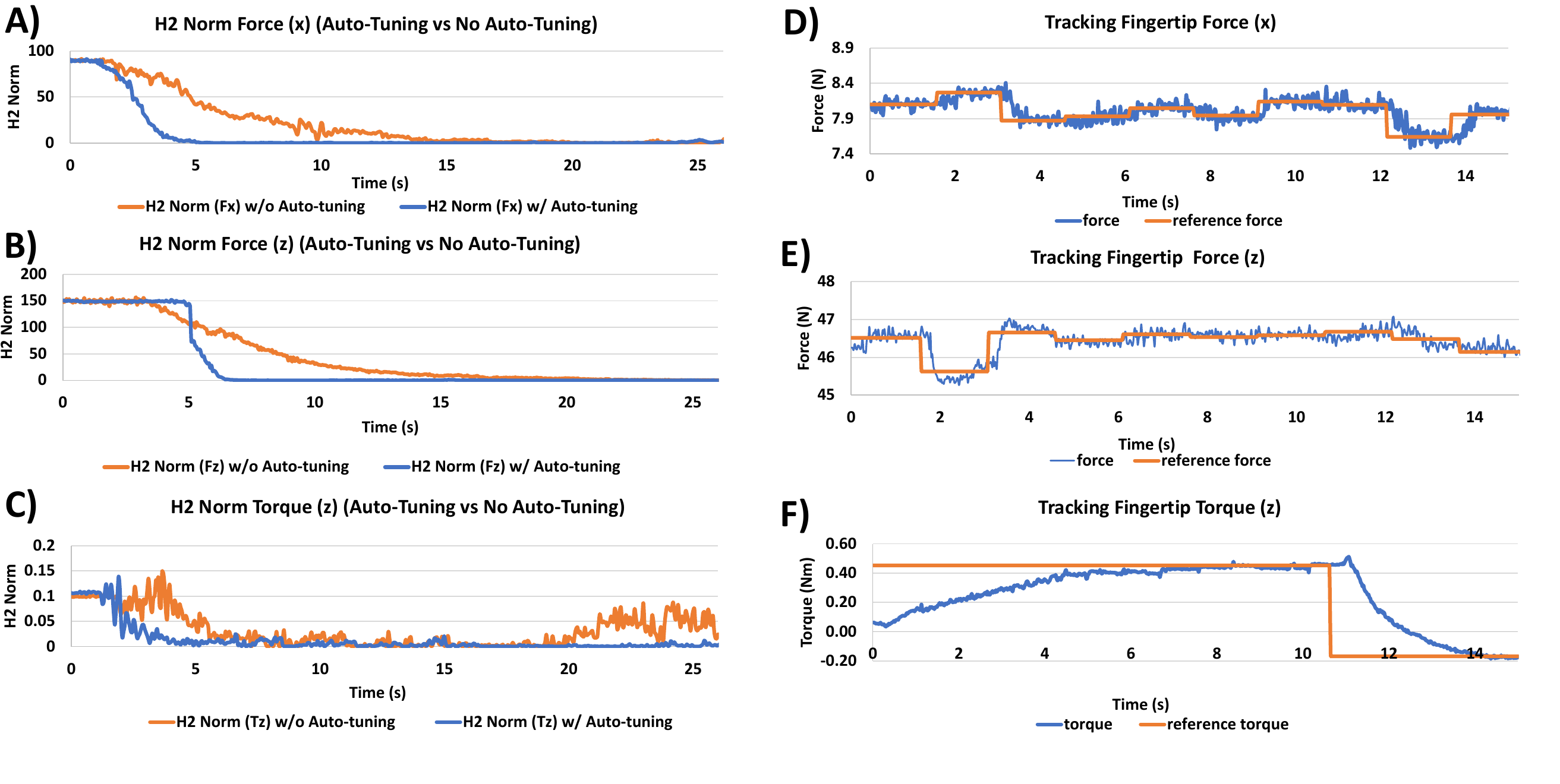}
    \caption{\textbf{Results: Tracking Wrench during Climbing}. A Climbing wrench trajectory is tracked in ($\mathbf{A}$) - ($\mathbf{F}$) or ${f}_{1,2,x}$, ${f}_{1,z}$, and ${\tau}_{1,2,z}$. The experiment is depicted on the left of Figure (\ref{intro_pic}).}
    \label{results_2}
\end{figure*}
\begin{figure}[!t]
    \centering
    \includegraphics[width=1\columnwidth]{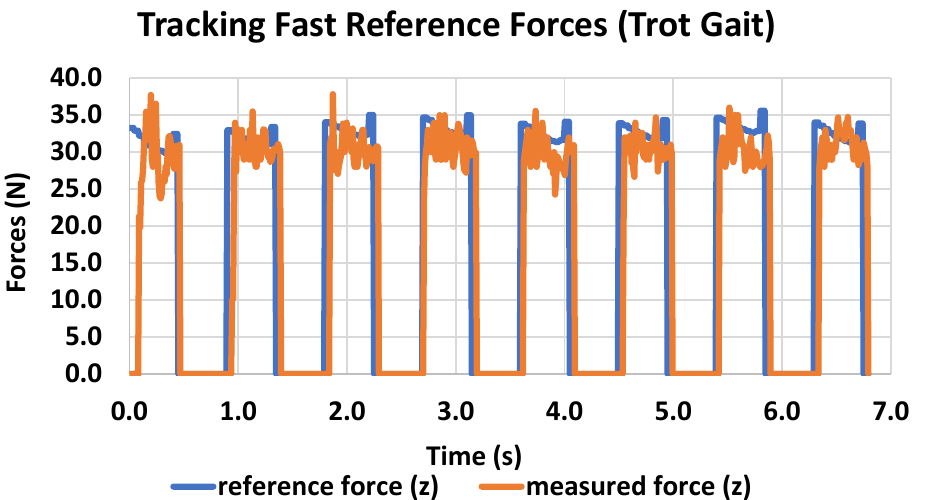}
    \caption{\textbf{Results: Tracking Fast Reference Forces from an MPC.} An MPC is used to generate reference forces during a trot gait (blue). The actual force (tracked by our adaptive admittance controller), is measured by FT sensors attached to the feet of the robot. The experiment is depicted on the bottom right of Figure (\ref{intro_pic}).
    }
    \label{results_3}
\end{figure}

\subsection{Training objectives}
\label{training_objective}
As described in Sec. \ref{methods:adaptive}, based on a combination of different $\boldsymbol{\theta}^{i}_{k,t}$ sigma points, which represent different sets of admittance controller gains and spring constants, and simulating them using our proposed model propagation as estimated by \eqref{eq:sys_dyn}, we then evaluate the performance of these sigma points using the evaluation function $\mathbf{h}(\boldsymbol{\theta}_{k,t}^{i})$. The objective is to drive our control parameters towards minimizing the difference between the outputs of our evaluation function with a user-defined desired output, specified by $\mathbf{y}_{k,t}^{\rm des}$ from \eqref{eq:learning_law}. In this section, we will now explicitly state our control parameters, $\boldsymbol{\theta}_{k,t}^i$, for each fingertip $k$, and describe our evaluation function $\mathbf{h}_{k,t}(\boldsymbol{\theta}^{i})$, which can be summarized as: 
\begin{equation}
    \boldsymbol{\theta}_{k,t}^i=
    \begin{bmatrix} 
    \mathbf{M}_{d}({\boldsymbol{\theta}^{i}_{M_{d}}})
    \\
    \mathbf{D}_{d}({\boldsymbol{\theta}^{i}_{D_{d}}})
    \\
    \mathbf{K}_{f}({\boldsymbol{\theta}^{i}_{K_{f}}})
    \\
    \mathbf{K}_{p}({\boldsymbol{\theta}^{i}_{p}})
    \\
    \mathbf{K}_{\Theta}({\boldsymbol{\theta}^{i}_{\Theta}})
    \end{bmatrix},
    \mathbf{h}_{k,t}(\boldsymbol{\theta}^i)=
    \begin{bmatrix}
    \mathbf{h}_{\rm ref}({\boldsymbol{\theta}}^{i})
    \\
    \mathbf{h}_{\rm spring}({\boldsymbol{\theta}}^{i})
    \\
    {h}_{\rm forces}({\boldsymbol{\theta}}^{i})
    \\
    {h}_{\rm const}({\boldsymbol{\theta}}^{i})
    \\
    \end{bmatrix}
    \end{equation}
\begin{align*}
    \mathbf{y}_{k,t}^{\rm des}=
    \begin{bmatrix}
    \mathbf{y}_{\rm ref}^{\rm des}
    \\
    \mathbf{y}_{\rm spring}^{\rm des}
    \\
    {y}_{\rm forces}^{\rm des}
    \\
    {y}_{\rm const}^{\rm des}
    \\
    \end{bmatrix}
\end{align*}
where the control parameters, $\boldsymbol{\theta}_{k,t}^i$ $\in\mathbb{R}^{24k}$, include the gains of the admittance controller, $\mathbf{M}_{d}$, $\mathbf{D}_{d}$, and $\mathbf{K}_{f}$, which in total includes 18 gains to be tuned per fingertip, and the spring constants $\mathbf{K}_{p}$ and $\mathbf{K_{\Theta}}$, which in total includes 6 constants to be tuned per $x$, $y$ and $z$ components. Note, we do not tune $\mathbf{K}_{d}$ in \eqref{eq:admittance_controller_inv} because in this work we are primarily interested in tracking force not position, i.e., the gripper must be in contact and we set $\mathbf{K}_{d}$ to be zero. The evaluation function $\mathbf{h}_{k,t}(\boldsymbol{\theta}^{i})\in\mathbb{R}^{14k}$ contains 4 main components, including the cost of following a reference wrench trajectory ($\mathbf{h}_{\rm ref}\in\mathbb{R}^{6k}$), the cost of estimating the spring constants ($\mathbf{h}_{\rm spring}\in\mathbb{R}^{6k}$), the cost of ensuring that the control output generates forces that do not cause slipping (${h}_{\rm forces}\in\mathbb{R}^{k}$), and the cost of satisfying kinematic and controller constraints (${h}_{\rm const}\in\mathbb{R}^{k}$). The costs of each will now be described in the following subsections and also include the corresponding desired values, represented by $\mathbf{y}_{k,t}^{\rm des}$ (which must be the same size and format as $\mathbf{h}_{k,t}(\boldsymbol{\theta}^{i})$).
\subsubsection{Reference trajectory}
\label{reference_trajectory}
One of the main goals for auto-tuning the admittance controller is to tune the gains such that a reference wrench trajectory, which was calculated using the methods found in \cite{risk_aware}, can be closely followed. To do so, for fingertip $k$ at timestep $t$, where $k,t$ notation are omitted for simplicity, we let $\mathbf{h}_{\rm ref}(\boldsymbol{\theta}^{i}) = [\mathcal{W}(\boldsymbol{\theta}^{i})^{G}]$ or specifically, $\mathbf{h}_{\rm ref}(\boldsymbol{\theta}^{i})=[\mathbf{f}(\boldsymbol{\theta}^{i})^{G\top},\boldsymbol{\tau}(\boldsymbol{\theta}^{i})^{G\top}]^\top$, which represent the fingertip force and torque values based on the model propagation using \eqref{eq:sys_dyn} to evaluate $\mathcal{W}(\boldsymbol{\theta}^{i})^{G}$. Note, that both are functions of the sigma points or $\boldsymbol{\theta}^{i}$ which provide the current gains of the admittance controller in \eqref{eq:admittance_controller_inv}. If we then let our desired parameters of \eqref{eq:learning_law} or $\mathbf{y}_{\rm ref}^{\rm des}=[\mathcal{W}_{\rm ref}^{G}]$, where $\mathcal{W}_{\rm ref}^{G}$ is received directly from the reference trajectory from \eqref{eq:admittance_controller_inv}, then the training objective is to produce control outputs that follow this trajectory as closely as possible. 
\subsubsection{Spring evaluation}
\label{spring_eval}
One of the challenges of the model propagation described in \eqref{eq:sys_dyn} is in propagating the fingertip wrenches from timestep $t-1$ to $t$. The approach used in this work is to assume a spring model, where we estimate the wrench at timestep $t$ by propagating the fingertip position and orientation instead, as shown in \eqref{eq:force_model} and \eqref{eq:torque_model}. However, to use this model we must know the spring constants $\mathbf{K}_{p}$ and $\mathbf{K}_{\Theta}$ for the $x$, $y$, and $z$ displacements, which requires extensive system identification methods. To automate this process, we can also use our auto-tuning method to evaluate $\mathbf{K}_{p}$ and $\mathbf{K}_{\Theta}$ during system operation directly. Thus, we have $\mathbf{h}_{\rm spring}(\boldsymbol{\theta}^{i}) = [\mathbf{K}_{p}(\boldsymbol{\theta}^{i})\Delta \mathbf{p}(\boldsymbol{\theta}^{i})^{G\top},\mathbf{K}_{\Theta}(\boldsymbol{\theta}^{i})\Delta \boldsymbol{\Theta}(\boldsymbol{\theta}^{i})^{G\top}]^\top$, where the values of the sigma points $\boldsymbol{\theta}^{i}$ are equivalent to the spring constant $\mathbf{K}_{p}$ and $\mathbf{K}_{\Theta}$ itself. Our desired parameter then becomes $\mathbf{y}_{\rm spring}^{\rm des} = [\mathbf{f}_{\rm meas}^{G\top},\boldsymbol{\tau}_{\rm meas}^{G\top}]^\top$, where $\mathbf{f}_{\rm meas}^{G}$ and $\boldsymbol{\tau}_{\rm meas}^{G}$ are the actual fingertip force and torque. By minimizing the difference between $\mathbf{h}_{\rm spring}(\boldsymbol{\theta}^{i})$ and $\mathbf{y}^{\rm des}_{\rm spring}$ (in \eqref{eq:learning_law}), the training objective of the auto-tuner is to find parameter $\boldsymbol{\theta}^{i}$ which best estimates the value of $\mathbf{K}_{p}$ and $\mathbf{K}_{\Theta}$ such that the spring model gets closer to the actual current values from (\ref{A_matrix}). Note, that by improving the model through updating the spring constants, we also improve the performance of the auto-tuner over time as the auto-tuning method relies on estimating the model mismatch in \eqref{eq:sys_dyn} for propagation.
\subsubsection{Force limits}
\label{force_eval}
During the evaluation of our costs, or $\mathbf{h}_{k,t},({\boldsymbol{\theta}^{i}})$, which requires simulating each sigma point using our dynamics model from timestep $t-1$ to $t$, it is possible that some combination of controller gains may propagate forces that are near or at the boundary of slipping (i.e., slipping between the fingertip contact point and its environment). As described previously, and written in more detail in \cite{kao_contact_2008}, we assume a contact model which has a finite contact patch that includes frictional and normal forces, and a torsional moment with respect to the contact normal. To prevent control outputs that cause slippage, we can make use of \eqref{eq:rotation_torque_bounds}: 
\begin{align}
    \label{eq:rotation_torque_bounds}
     -{\lambda} {f}_{k,z}^{G} \leq {\tau}^{G}_{k,z} \leq {\lambda}{f}_{k,z}^{G}
\end{align}
where the fingertip is least likely to slip when ${\tau}_{k,z}^{G}$ is between $-{\lambda}{f}^{G}_{k,z}$ and ${\lambda}{f}^{G}_{k,z}$. This behavior can be achieved by setting an arbitrary high cost (say $\delta$) when \eqref{eq:rotation_torque_bounds} is not satisfied, and if \eqref{eq:rotation_torque_bounds} is satisfied, we can set the cost to be a function of how  $\lq$far' away the system is from slipping. Thus, we can let ${h}_{\rm forces}(\boldsymbol{\theta}^{i})=[{\delta}]$ if \eqref{eq:rotation_torque_bounds} is not satisfied, and if it is satisfied, we can let ${h}_{\rm forces}(\boldsymbol{\theta}^{i})=[\frac{1}{|{\lambda} {f}_{k,z}^{G}|-|{\tau}_{k,z}^{G}|}]$. In both cases, our desired parameter is ${y}_{\rm forces}^{\rm des}=[{0}]$




\subsubsection{Kinematic and controller constraints}
\label{kinem_eval}
As we propagate our admittance control output $\mathbf{x}$ based on different values for $\boldsymbol{\theta}^{i}$ during our propagation from $t-1$ to $t$, we may get unreasonable values for $\mathbf{x}$, due to $\lq$bad' choices of $\boldsymbol{\theta}^{i}$, that are kinematically infeasible or outside the workspace of the robot causing singularity, or obtain controller gains which are not positive semi-definite. A simple solution to avoid our auto-tuning method to choose these $\boldsymbol{\theta}^{i}$ values, is to impose a large cost, which we arbitrarily term as $\zeta$, when $\mathbf{x}$ is infeasible or violate the semi-definite property. Thus, we can write that if $\boldsymbol{\theta}^{i}$ are chosen such that it causes a kinematic infeasibility or violates the positive semi-definite property then $\mathbf{h}_{\rm const}(\boldsymbol{\theta}^{i})=[\zeta]$, and if the solution does not cause violations, then $\mathbf{h}_{\rm const}(\boldsymbol{\theta}^{i})=0$, where our desired value $y_{\rm const}^{\rm des}=[0]$.

\section{Experimental Validation}
\label{results}
\label{implementation}
We implement our controller on the SCALER \cite{tanaka2021underactuated} quadruped climbing robot. SCALER weighs 9.6 kg, and has two configurations, a walking configuration used to track ground reaction forces from an MPC \cite{forceMPC}, 3 DoF per leg, and a climbing configuration to track fingertip wrenches, 6 DoF per leg in addition to a 1 DoF two-finger gripper, which totals 7 DoF per limb \cite{tanaka2021underactuated}. At the wrist of the robot, or endpoint of 3 DoF or 7 DoF configuration, we use BOTA's force/torque sensor \cite{bota}, and have 1-axis strain gauges at each fingertip of our gripper, where actuation for grasping is done using a linear actuator. We demonstrate the efficacy of auto-tuning our gains using the H2 norm defined as  $||\mathcal{W}_{\rm meas} - \mathcal{W}_{\rm ref}||_{H_2}$ (defined as a scalar value). Thus, a decrease in the H2 norm indicates better tracking of the reference wrench trajectory. Overall, we show our results through three different experiments as shown in Figures \ref{results_1}, \ref{results_2} and \ref{results_3}. In Sec. \ref{experiment1}, we discuss the tuning process for grasping, and apply our method toward tracking grasping force on objects with different degrees of compliance. In Sec. \ref{experiment2}, we discuss the tuning and then tracking of not only forces, but also torques, which is necessary for free-climbing tasks. In Sec. \ref{experiment3}, to show how our admittance controller can handle tracking reference forces that rapidly changes, we apply our controller for tracking the ground reaction force outputs of an MPC controller during a fast moving trot gait. For \ref{experiment1} and \ref{experiment2}, our baseline or no auto-tuning, is an admittance controller that has been hand-tuned to the best of our abilities to achieve convergence according to \cite{tune_admitt}. We do not show a baseline for \ref{experiment3}, as we were not able to hand-tune the gains of the admittance controller that could successfully track or converge for fast moving reference forces. Finally, we note that we ran the admittance controller and auto-tuner at 100 Hz, and applied the auto-tuning procedure for one fingertip only. 

\subsection{Experiment 1: Tracking grasping force}
\label{experiment1}
As shown in ($\mathbf{A}$)-($\mathbf{C}$) in Fig. \ref{results_1}, we were able to track the grasping force of a stiff object, or bouldering hold, as well as compliant objects, or paper towel and a soft ball. We show the tuning results in ($\mathbf{D}$), where we see a lower H2 norm with auto-tuning (blue) compared to not using auto-tuning (orange), using the bouldering hold as seen in ($\mathbf{A}$) as our example. Additionally, without auto-tuning the system diverges and becomes unstable as shown by the large peaks. In ($\mathbf{E}$) we show how the auto-tuner updates the admittance controller gains, here we only update $M_{d}$ and $D_{d}$, which show that both gains initialized at 0.1 converge to a constant value once the controller becomes stable or follows the reference trajectory. We also show in $\mathbf{(F)}$ how the $K_{p}$ spring constant, here just a single value, changes as the gripper is increasing its grasping force, where eventually it reaches a very high value as the gripper's change in position decreases with respect to increasing grasping force. Lastly, we note that the auto-tuner only needed 0.8 seconds to converge to optimal gains for this test.

\subsection{Experiment 2: Tracking wrenches for free-climbing}
\label{experiment2}
To auto-tune other force/torque components, we update $\mathbf{M}_d$, $\mathbf{D}_d$, $\mathbf{K}_f$ and spring constants by hanging the robot on a bar and having it grasp a bouldering hold while following a constant reference wrench trajectory. In total, we tune the following components of wrench: $f_{1,2,x}^{G}$, $f_{1,2,z}^{G}$, $\tau_{1,2,x}^{G}$, $\tau_{1,2,y}^{G}$, and $\tau_{1,2,z}^{G}$. ($\mathbf{A}$)-($\mathbf{C}$) of Fig. \ref{results_2} shows that with auto-tuning we demonstrate quicker convergence to our reference compared to without auto-tuning for $f_{1,2,x}^{G}$, $f_{1,2,z}^{G}$, and $\tau_{1,2,z}^{G}$ as an example. We note that with and without auto-tuning the H2 norm eventually converged for the case of tracking force, likely as the initialized gains were already sufficiently stable. However, without auto-tuning, the H2 norm could not converge for tracking the additional torque (see ($\mathbf{C}$) of Fig. \ref{results_2}), $\tau_{1,2,z}^{G}$, while convergence occurred using auto-tuning. Additionally, the auto-tuner took about 3 seconds to converge for fingertip force in the $x$ direction, 7 seconds for force in the $z$ direction, and 5 seconds for torque in the $z$ direction. 

After using auto-tuning to find optimal gains, we use these updated gains to track a changing wrench trajectory, namely shear force, grasping force with fingertip normal force offset as described in Sec. \ref{adm_implementation}, and additional torque due to rotational friction along the yaw axis, or $f_{1,2,x}^{G}$, $f_{1,z}^{G}$, $f_{2,z}^{G}$, and $\tau_{1,2,z}^{G}$ simultaneously (results for $f_{1,z}^{G}$ are shown for brevity). The results of tracking this wrench is demonstrated in ($\mathbf{D}$)-($\mathbf{F}$) in Fig. \ref{results_2} for the robotic free-climbing task.

\subsection{Experiment 3: Tracking fast reference forces}
\label{experiment3}
Lastly, we demonstrate the speed of our admittance controller when applied to a single point contact robot during a trot gait as shown in Fig. \ref{results_3}. Output ground reaction forces from a point contact ($f_{z}^{G}$) are produced by an MPC controller, formulated identically to \cite{forceMPC}, and we show a step response of $\approx$ 0.03 seconds across 8 steps of a trot gait (i.e., it took about 0.03 seconds for the admittance controller to track the force profile per step). We note that while the step response was adequate for climbing and even for locomotion tasks, the MPC only needs to run every $\approx$ 0.03 seconds, faster response rates may be achievable with motors that have a lower gear ratio, as our motors have a high gear ratio.

\section{Conclusion}
We demonstrated an auto-calibrating admittance controller that could track wrench trajectories during locomotion and manipulation tasks. We show that we could successfully track the additional torque due to rotational friction simultaneously with other force components for an extreme manipulation case trajectory - namely for robotic wall-climbing. In future work, we aim to use the auto-tuner to derive various spring constants for different surfaces, and examine more complex training objectives such as recovering from slipping.
\bibliographystyle{IEEEtran}
\bibliography{bibliography}

\end{document}